\def\year{2020}\relax

\documentclass[letterpaper]{article} 

\usepackage{aaai20}  

\usepackage{times}  
\usepackage{helvet} 
\usepackage{courier}  
\usepackage[hyphens]{url}  
\usepackage{graphicx} 
\usepackage{graphicx}  
\usepackage{ifxetex}

\usepackage{refcount}

\ifxetex
 \usepackage{fontspec}
\else
 \usepackage[utf8]{inputenc}
\fi

\newcommand{\citet}[1]{\citeauthor{#1} \shortcite{#1}}
\newcommand{\citep}{\cite}

\usepackage{amsmath,amsfonts,bm}

\def\eqref#1{equation~\ref{#1}}

\def\1{\bm{1}}

\def\rva{{\mathbf{a}}}

\DeclareMathAlphabet{\mathsfit}{\encodingdefault}{\sfdefault}{m}{sl}
\SetMathAlphabet{\mathsfit}{bold}{\encodingdefault}{\sfdefault}{bx}{n}

\def\gN{{\mathcal{N}}}
\def\gO{{\mathcal{O}}}
\def\gP{{\mathcal{P}}}

\def\sR{{\mathbb{R}}}

\def\emR{{R}}

\def\emV{{V}}
\def\emW{{W}}

\newcommand{\softmax}{\mathrm{softmax}}

\usepackage{graphicx}

\usepackage{subcaption}

\usepackage{enumitem}

\usepackage{xcolor}
\definecolor{nice-red}{HTML}{E41A1C}
\colorlet{dark-red}{nice-red!80!black}
\definecolor{nice-orange}{HTML}{FF7F00}
\colorlet{dark-orange}{orange!85!black}
\definecolor{nice-yellow}{HTML}{FFC020}
\definecolor{nice-green}{HTML}{4DAF4A}
\definecolor{nice-blue}{HTML}{377EB8}
\definecolor{nice-purple}{HTML}{984EA3}

\usepackage{url}            
\usepackage{booktabs}       
\usepackage{microtype}      

\usepackage[xindy]{glossaries}

\usepackage[]{todonotes}

\usepackage{bm}
\usepackage[capitalise]{cleveref}

\crefformat{equation}{Eq.~#2#1#3}
\Crefformat{equation}{Equation~#2#1#3}

\crefrangeformat{equation}{Eqs.~#3#1#4 to~#5#2#6}
\Crefrangeformat{equation}{Equations~#3#1#4 to~#5#2#6}
\crefmultiformat{equation}{Eqs.~#2#1#3}%
{ and~#2#1#3}{, #2#1#3}{ and~#2#1#3}
\Crefmultiformat{equation}{Equations~#2#1#3}%
{ and~#2#1#3}{, #2#1#3}{ and~#2#1#3}

\usepackage{tikz}
\usetikzlibrary{fit,shapes.misc}

\newcommand\marktopleft[1]{%
    \tikz[overlay,remember picture] 
        \node (marker-#1-a) at (0,1.5ex) {};%
}
\newcommand\markbottomright[1]{%
    \tikz[overlay,remember picture] 
        \node (marker-#1-b) at (0,0) {};%
    \tikz[overlay,remember picture,thick,inner sep=3pt]
        \node[draw,rounded rectangle,fit=(marker-#1-a.center) (marker-#1-b.center)] {};%
}

\usetikzlibrary{calc,trees,positioning,arrows,chains,shapes.geometric,%
  decorations.pathreplacing,decorations.pathmorphing,shapes,%
  matrix,shapes.symbols,fit,decorations,arrows.meta}

\usepackage{amsmath}
\usepackage{amsthm}

\theoremstyle{definition}

\usepackage{lipsum}
\usepackage{multirow}

\usepackage{amssymb}
\usepackage{soul}

\usepackage{quoting}
\quotingsetup{vskip=5pt,indentfirst=false,leftmargin=5pt,rightmargin=5pt}

\usepackage{paralist}
\usepackage{tabularx}
\usepackage{lscape}
\usepackage[misc]{ifsym}

\newcolumntype{b}{X}
\newcolumntype{s}{>{\hsize=.2\hsize}X}
\newcolumntype{m}{>{\hsize=.4\hsize}X}
\newcolumntype{z}{>{\hsize=.8\hsize}X}

\usepackage{rotating}

\usepackage{array}
\newcolumntype{L}[1]{>{\raggedright\let\newline\\\arraybackslash\hspace{0pt}}m{#1}}
\newcolumntype{C}[1]{>{\centering\let\newline\\\arraybackslash\hspace{0pt}}m{#1}}
\newcolumntype{R}[1]{>{\raggedleft\let\newline\\\arraybackslash\hspace{0pt}}m{#1}}

\newacronym{AKBC}{AKBC}{Automated Knowledge Base Construction}
\newacronym{AUC}{AUC}{Area Under Curve}

\newacronym{AUC-PR}{AUC-PR}{Area Under the Precision-Recall Curve}
\newacronym{BPR}{BPR}{Bayesian Personalized Ranking}

\newacronym{CNF}{CNF}{Conjunctive Normal Form}
\newacronym{CNN}{CNN}{Convolutional Neural Network}
\newacronym{CWA}{CWA}{Closed World Assumption}

\newacronym{FOIL}{FOIL}{First Order Inductive Learner}

\newacronym{GPCA}{GPCA}{Generalized Principal Component Analysis}
\newacronym{GPU}{GPU}{Graphics Processing Unit}

\newacronym{ILP}{ILP}{Inductive Logic Programming}
\newacronym{dILP}{dILP}{Differentiable Inductive Logic Programming}

\newacronym{AI}{AI}{Artificial Intelligence}
\newacronym{MR}{MR}{Machine Reading}
\newacronym{NLU}{NLU}{Natural Language Understanding}

\newacronym{KB}{KB}{Knowledge Base}
\newacronym{KG}{KG}{Knowledge Graph}

\newacronym[longplural={Long Short-Term Memories}]{LSTM}{LSTM}{long short-term memory}

\newacronym{NLP}{NLP}{Natural Language Processing}
\newacronym{NLI}{NLI}{Natural Language Inference}

\newacronym{NTP}{NTP}{Neural Theorem Prover}

\newacronym{SOTA}{SOTA}{state-of-the-art}

\newacronym{NaNTP}{GNTP}{Greedy NTP}

\newacronym{NTN}{NTN}{Neural Tensor Network}

\newacronym{MAP}{MAP}{Mean Average Precision}
\newacronym{MLP}{MLP}{Multi-layer Perceptron}
\newacronym{MRR}{MRR}{Mean Reciprocal Rank}

\newacronym{OpenIE}{OpenIE}{Open Information Extraction}

\newacronym{PCA}{PCA}{Principal Component Analysis}
\newacronym{PRA}{PRA}{Path Ranking Algorithm}
\newacronym{ProPPR}{ProPPR}{Programming with Personalized PageRank}

\newacronym{RBF}{RBF}{Radial Basis Function}
\newacronym{ROC}{ROC}{Receiver Operating Characteristic}
\newacronym{RNN}{RNN}{Recurrent Neural Network}
\newacronym{RTE}{RTE}{Recognizing Textual Entailment}

\newacronym{SGD}{SGD}{Stochastic Gradient Descent}
\newacronym{SNLI}{SNLI}{Stanford Natural Language Inference}

\newacronym{ANNS}{ANNS}{Approximate Nearest Neighbour Search}
\newacronym{NNS}{NNS}{Nearest Neighbour Search}

\newacronym{LSH}{LSH}{Locality-Sensitive Hashing}
\newacronym{PQ}{PQ}{Product Quantization}
\newacronym{PG}{PG}{Proximity Graph}
\newacronym{HNSW}{HNSW}{Hierarchical Navigable Small World}

\newacronym{MoE}{MoE}{Mixture of Experts}

\newacronym{UMLS}{UMLS}{Unified Medical Language System}

\newacronym{FAISS}{FAISS}{Facebook AI Similarity Search}

\newcommand{\eg}{\emph{e.g.}}
\newcommand{\ie}{\emph{i.e.}}

\newcommand{\kb}{\mathfrak{K}}
\newcommand{\pkb}{\mathfrak{P}}
\newcommand{\state}{S}

\newcommand{\krnl}{k}

\renewcommand{\emptyset}{\varnothing}

\newcommand{\fun}[1]{\text{#1}}

\let\set\dom 
\newcommand{\lss}[1]{\mathbb{#1}}
\let\ls\textsc
\let\lst\ls

\newcommand{\xs}[1]{\bm{[} #1 \bm{]}}

\renewcommand\vec[1]{{\bm{#1}}}

\makeatletter
\newcommand*\bdot{\mathpalette\bdot@{.5}}
\newcommand*\bdot@[2]{\mathbin{\vcenter{\hbox{\scalebox{#2}{$\m@th#1\bullet$}}}}}
\makeatother

\def\params{{\vec{\theta}}}

\newcommand{\rel}[1]{\verb~#1~}
\newcommand{\const}[1]{\textsc{#1}}
\let\ent\const
\newcommand{\var}[1]{\textsc{#1}}
\def\lif{\ \text{:--}\ }

\let\subs\psi

\newcommand{\pred}[1]{\text{$\verb~#1~$}}

\def \Real {\mathbb{R}}

\let\success\rho

\let\todonote\todo

\colorlet{fixme}{red!85!black}
\colorlet{fixme-bright}{fixme!25}

\colorlet{todo}{orange!85!black}
\colorlet{todo-bright}{todo!25}

\colorlet{review}{nice-yellow!50!nice-orange}
\colorlet{review-bright}{review!25}

\colorlet{maybe}{nice-yellow}
\colorlet{maybe-bright}{maybe!25}

\colorlet{toref}{purple!70!black}
\colorlet{toref-bright}{toref!25}

\colorlet{info}{green!50!black}
\colorlet{info-bright}{info!25}

\colorlet{cut}{black!60}
\colorlet{cut-bright}{cut!25}

\colorlet{extend}{blue!85!black}
\colorlet{extend-bright}{extend!25}

\colorlet{discuss}{nice-red!50!black}
\colorlet{discuss-bright}{discuss!25}

\renewcommand{\todo}[1]{\todonote[linecolor=todo, backgroundcolor=todo-bright, bordercolor=todo]{\thesection{} {\bf\color{todo}TODO} #1}{}}

\newcommand{\info}[1]{}
\newcommand{\hlinfo}[2]{}

\newcommand{\maybe}[1]{}
\newcommand{\hlmaybe}[2]{}

\renewcommand{\const}[1]{{\color{black}\textsc{#1}}}

\let\ent\const

\renewcommand{\var}[1]{{\color{black}\textsc{#1}}}

\newcommand{\module}[1]{{\verb~#1~}}
\newcommand{\datadom}[1]{{#1}}
\newcommand{\tensordom}[1]{{#1}}

\usepackage{makecell}
\usepackage{tablefootnote}
\usepackage{colortbl}

\quotingsetup{vskip=1pt,leftmargin=10pt,rightmargin=10pt}

\title{Differentiable Reasoning on Large Knowledge Bases and Natural Language}

\usepackage[capitalise]{cleveref}

\newcommand{\ucl}{$^1$}
\newcommand{\fair}{$^2$}
\newcommand{\uclfair}{$^{1,2}$}

\author{Pasquale Minervini\thanks{Equal contribution}\thanks{Corresponding author}\ucl{} \quad Matko Bošnjak\footnotemark[1]\thanks{Now at DeepMind}\ucl{}, \\ \bf \Large Tim Rocktäschel\uclfair{} \quad Sebastian Riedel\uclfair{} \quad Edward Grefenstette\uclfair{} \\
  \ucl{}UCL Centre for Artificial Intelligence, University College London \\
  \fair{}Facebook AI Research \\
  \texttt{\{p.minervini,m.bosnjak,t.rocktaschel,s.riedel,e.grefenstette\}@cs.ucl.ac.uk} \\
}

\begin{document}

\nocopyright

\maketitle

\begin{abstract}
Reasoning with knowledge expressed in natural language and \glspl{KB} is a major challenge for Artificial Intelligence, with applications in machine reading, dialogue, and question answering.
General neural architectures that jointly learn representations and transformations of text are very data-inefficient, and it is hard to analyse their reasoning process.
These issues are addressed by end-to-end differentiable reasoning systems such as \glspl{NTP}, although they can only be used with small-scale symbolic \glspl{KB}.
In this paper we first propose \glspl{NaNTP}, an extension to \glspl{NTP} addressing their complexity and scalability limitations, thus making them applicable to real-world datasets.
This result is achieved by dynamically constructing the computation graph of \glspl{NTP} and including only the most promising proof paths during inference, thus obtaining orders of magnitude more efficient models~\footnote{Source code, datasets, and supplementary material are available online at \url{https://github.com/uclnlp/gntp}.}.
Then, we propose a novel approach for jointly reasoning over \glspl{KB} and textual mentions, by  embedding logic facts and natural language sentences in a shared embedding space.
We show that \glspl{NaNTP} perform on par with \glspl{NTP} at a fraction of their cost while achieving competitive link prediction results on large datasets, providing explanations for predictions, and inducing interpretable models.
\end{abstract}
\section{Introduction} \label{sec:introduction}
The main focus of Artificial Intelligence is building systems that exhibit intelligent behaviour~\citep{DBLP:journals/ai/Levesque14}. 
Notably, \gls{NLU} and \gls{MR} aim at building models and systems with the ability to read text, extract meaningful knowledge, and reason with it~\citep{etzioni2006machine,hermann2015teaching,DBLP:journals/corr/WestonBCM15,DBLP:conf/eacl/McCallumNDB17}.
This ability facilitates both the synthesis of new knowledge and the possibility to verify and update a given assertion.
\begin{figure*}[t]
\centering
\includegraphics[width=\textwidth]{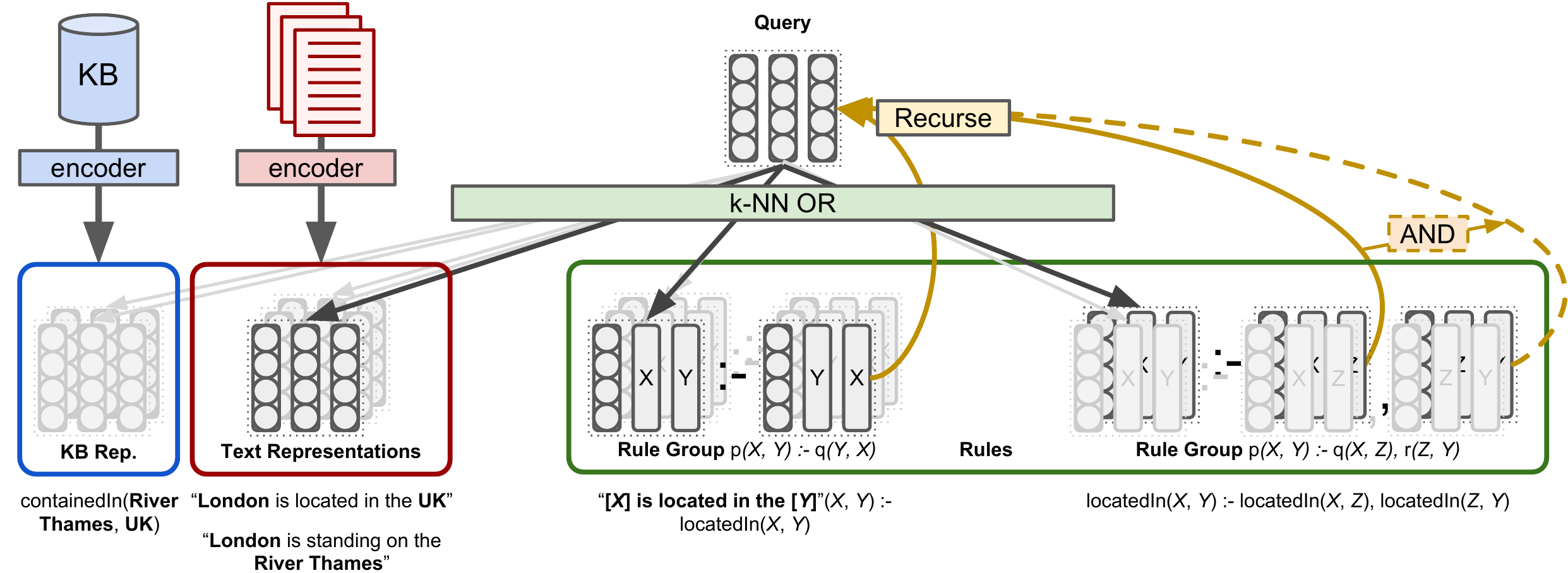}
\caption{Overall architecture of \glspl{NaNTP}. The two main contributions lie in i) the significantly faster inference mechanism, sped up by the k-NN OR component, and ii) the text encoder.}
\label{fig:overview}
\end{figure*}
Traditionally, automated reasoning applied to text requires natural language processing tools that compile it into the structured form of a \gls{KB}~\citep{DBLP:journals/corr/abs-1806-05599}.
However, the compiled \glspl{KB} tend to be incomplete, ambiguous, and noisy, impairing the application of standard deductive reasoners~\citep{DBLP:conf/ijcai/HuangHT05}.
A rich and broad literature in \gls{MR} has approached this problem within a variety of frameworks, including Natural Logic~\citep{maccartney2007natural}, Semantic Parsing~\citep{bos2008wide}, Natural Language Inference and Recognising Textual
Entailment~\citep{Fyodorov-etal:2000,bowman2015large}, and Question Answering~\citep{hermann2015teaching}.
Nonetheless, such methods suffer from several limitations.
They rely on significant amounts of annotated data to suitably approximate the implicit distribution from which the data is drawn.
In practice, this makes them unable to generalise well in the absence of a sufficient quantity of training data or appropriate priors on model parameters~\citep{evans2018learning}.
Orthogonally, even when accurate, such methods cannot explain given predictions~\citep{DBLP:journals/cacm/Lipton18}.
A promising strategy for overcoming these issues consists of combining \emph{neural models} and \emph{symbolic reasoning}, given their complementary strengths and weaknesses~\citep{DBLP:conf/aaaiss/GarcezBRFHIKLMS15,rocktaschel2017end,DBLP:conf/nips/YangYC17,evans2018learning,DBLP:conf/acl/WeberMMLR19}.
While symbolic models can generalise well from a small number of examples, they are brittle and prone to failure when the observations are noisy or ambiguous, or when the properties of the domain are unknown or hard to formalise, all of which being the case for natural language~\citep{DBLP:conf/ilp/RaedtK08,GARNELO201917}.
Contrarily, neural models are robust to noise and ambiguity but not easily interpretable, making them unable to provide explanations or incorporating background knowledge~\citep{DBLP:journals/corr/abs-1802-01933}.
Recent work in neuro-symbolic systems has made progress towards end-to-end differentiable reasoning models that can be trained via backpropagation while maintaining interpretability and generalisation, thereby inheriting the best of both worlds.
Among such systems, \glspl{NTP}~\citep{rocktaschel2017end,DBLP:journals/corr/abs-1807-08204} are end-to-end differentiable deductive reasoners based on Prolog's backward chaining algorithm, where discrete unification between atoms is replaced by a differentiable operator computing the similarities between their embedding representations.
\glspl{NTP} are especially interesting since they allow learning \emph{interpretable rules} from data, by back-propagating the prediction errors to the rule representations.
Furthermore, the proving process in \glspl{NTP} is \emph{explainable} -- the proof path associated with the largest proof score denotes which rules and facts are used in the reasoning process.
However, \glspl{NTP} have only been successfully applied to learning tasks involving very small datasets, since their computational complexity makes them unusable on larger, real-world \glspl{KB}.
Furthermore, most human knowledge is not available in \glspl{KB}, but in natural language texts which are difficult to reason over automatically.
In this paper we address these issues by proposing: 
\begin{inparaenum}[\itshape i)\upshape]
\item two efficiency improvements for significantly reducing the time and space complexity of \glspl{NTP} by reducing the number of candidate proof paths and introducing an attention mechanism for rule induction, and
\item an extension of \glspl{NTP} towards natural language, jointly embedding predicates and textual surface patterns in a shared space by using an end-to-end differentiable reading component.
\end{inparaenum}
\section{End-to-end Differentiable Proving} \label{sec:ntp}
\glspl{NTP}~\citep{rocktaschel2017end} recursively build a neural network enumerating all the possible proof paths for proving a query (or \emph{goal}) on a given \gls{KB}, and aggregate all their proof scores via max pooling.
They do so by relying on three modules---a \emph{unification module}, which compares sub-symbolic representations of logic atoms, and mutually recursive \emph{or} and \emph{and modules}, which jointly enumerate all possible proof paths, before the final aggregation selects the highest-scoring one.
In the following, we briefly overview these modules, and the training process used for learning the model parameters from data.
We assume the existence of a function-free Datalog \gls{KB} $\kb$ containing \emph{ground facts} in the form $\xs{\rel{p}, \const{a}, \const{b}}$~\footnote{For consistency, we use the same notation as \citet{rocktaschel2017end}.}, representing the logical atom $\rel{p}(\const{a}, \const{b})$ where $\rel{p}$ is a relation type, and $\const{a}, \const{b}$ are its arguments.~\footnote{We consider binary predicates, without loss of generality.}
It also contains \emph{rules} in the form $\lst{H} \lif \lst{B}$ such as $\xs{\rel{p}, \var{X}, \var{Y}} \lif \xs{\xs{\rel{q}, \var{X}, \var{Z}}, \xs{\rel{r}, \var{Z}, \var{Y}}}$, denoting the rule $\rel{p}(\var{X}, \var{Y}) \lif \rel{q}(\var{X}, \var{Z}), \rel{r}(\var{Z}, \var{Y})$, meaning that $\rel{q}(\var{X}, \var{Z}), \rel{r}(\var{Z}, \var{Y})$ implies $\rel{p}(\var{X}, \var{Y})$, where $\var{X}, \var{Y}, \var{Z}$ are universally quantified variables.
\paragraph{Unification Module.}
In the backward chaining reasoning algorithm, \emph{unification} is the operator that matches two logic atoms, such as $\rel{locatedIn}(\const{london}, \const{uk})$ and $\rel{situatedIn}(\var{X}, \var{Y})$.
Discrete unification checks for equality between the elements composing the two atoms (\eg{} $\rel{locatedIn} \neq \rel{situatedIn}$), and binds variables to symbols via substitutions (\eg{} $\{ \var{X}/\const{london}, \var{Y}/\const{uk} \}$).
In \glspl{NTP}, unification matches two atoms by comparing their \emph{embedding representations} via a differentiable similarity function -- a Gaussian kernel -- which enables matching different symbols with similar semantics.
More formally, $\module{unify}_\params(\lst{H}, \lst{G}, \lst{S}) = \state'$ creates a neural network module that matches two atoms $\lst{H}$ and $\lst{G}$ by comparing their embedding vectors.
For instance, given a goal $\lst{G} = \xs{\rel{locatedIn}, \const{London}, \const{UK}}$, a fact $\lst{H} = \xs{\rel{situatedIn}, \var{X}, \var{Y}}$, and a proof state $\state = (\state_\subs, \state_\success)$ consisting of a set of substitutions $\state_\subs$ and a proof score $\state_\success$, the \module{unify} module compares the embedding representations of $\rel{locatedIn}$ and $\rel{situatedIn}$ with a Gaussian kernel $\krnl$, updates the variable binding substitution set $\state'_\subs = \state_\subs \cup \{ \var{X}/\const{London}, \var{Y}/\const{UK} \}$, and calculates the new proof score $\state'_\success = \min \left(\state_\success,  \krnl\left(\params_{\scriptsize\rel{locatedIn}:}, \params_{\scriptsize\rel{situatedIn}:}\right) \right)$
and proof state $\state' = (\state'_\subs, \state'_\success)$. 
\paragraph{OR Module.}
The \module{or} module computes the unification between a goal and all facts and rule heads in a \gls{KB}, and then recursively invokes the \module{and} module on the corresponding rule bodies.
Formally, for each rule $\lst{H} \lif \lst{B}$~\footnote{Facts are seen as rules with no body and variables, \ie{} $\lst{F} \lif \xs{}$.} in a \gls{KB} $\kb$, $\module{or}^{\kb}_\params(\lst{G}, d, \state)$ unifies the goal $\lst{G}$ with the rule head $\lst{H}$, and invokes the \module{and} module to prove atoms in the body $\lst{B}$, keeping track of the maximum proof depth $d$:
\begin{equation} \label{eq:or}
\begin{aligned}
\module{or}^{\kb}_\params( \lst{G}, d, \state) & = [ \state' \mid \lst{H} \lif \lst{B} \in \kb, \\
& \state' \in \module{and}^{\kb}_\params(\lst{B}, d, \module{unify}_\params(\lst{H}, \lst{G}, \state)) ]
\end{aligned}
\end{equation}
For example, given a goal $\lst{G} = [\rel{situatedIn}, \var{Q}, \const{UK}]$ and a rule $\lst{H} \lif \lst{B}$ with $\lst{H} = \xs{\rel{locatedIn}, \var{X}, \var{Y}}$ and $\lst{B} = \xs{\xs{\rel{locatedIn},\var{X},\var{Z}}, \xs{\rel{locatedIn}, \var{Z},\var{Y}}}$, the model would unify the goal $\lst{G}$ with the rule head $\lst{H}$, and invoke the \module{and} modules to prove the sub-goals in the rule body $\lst{B}$.
\paragraph{AND Module.}
The \module{and} module recursively proves a list of sub-goals in a rule body.
Given the first sub-goal $\ls{B}$ and the following sub-goals $\lss{B}$, the $\module{and}^{\kb}_\params(\ls{B}:\lss{B}, d, \state)$ module will substitute variables in \ls{B} with constants according to the substitutions in $\state$, and invoke the \module{or} module on $\ls{B}$.
The resulting state is used to prove the atoms in $\lss{B}$, by recursively invoking the \module{and} module:
\begin{equation} \label{eq:and}
\begin{aligned}
\module{and}^{\kb}_\params(\ls{B}:\lss{B}, d, \state) & = [ \state'' \mid d > 0, \\
& \state''\in\module{and}^{\kb}_\params(\lss{B}, d, \state'), \\
& \state' \in \module{or}^{\kb}_\params(\module{sub}(\ls{B}, \state_\subs), d - 1, \state) ]
\end{aligned}
\end{equation}
For example, when invoked on the rule body $\lst{B}$ of the example mentioned above, the \module{and} module will substitute variables with constants for the sub-goal $\xs{\rel{locatedIn}, \var{X}, \var{Z}}$ and invoke the \module{or} module, whose resulting state will be the basis of the next invocation of \module{and} module on $\xs{\rel{locatedIn}, \var{Z}, \var{Y}}$.
\paragraph{Proof Aggregation.}
After building a neural network that evaluates all the possible proof paths of a goal $\lst{G}$ on a \gls{KB} $\kb$, \glspl{NTP} select the proof path with the largest proof score:
\begin{equation} \label{eq:max}
\begin{aligned}
& \module{ntp}^{\kb}_\params(\lst{G}, d) = \max_{S} \state_\success \\
& \text{with} \quad \state \in \module{or}^{\kb}_\params(\lst{G}, d, (\emptyset, 1))
\end{aligned}
\end{equation}
where $d \in \mathbb{N}$ is a predefined maximum proof depth. The initial proof state is set to $(\emptyset, 1)$ corresponding to an empty substitution set and to a proof score of $1$.
\paragraph{Training.}
In \glspl{NTP}, embedding representations are learned by minimising a cross-entropy loss $\mathcal{L}^{\kb}(\params)$ on the final proof score, by iteratively masking facts in the \gls{KB} and trying to prove them using other available facts and rules.
Negative examples are obtained via a corruption process, denoted by $\fun{corrupt}(\cdot)$, by modifying the subject and object of triples in the \gls{KB}~\citep{DBLP:journals/pieee/Nickel0TG16}:
\begin{equation} \label{eq:loss}
\begin{aligned}
\mathcal{L}^{\kb}(\params) = & - \sum_{\lst{F} \lif \xs{} \in \kb} \log \module{ntp}^{\kb \setminus \lst{F}}_\params(\lst{F}, d) \\
& - \sum_{\tilde{\lst{F}} \sim \fun{corrupt}(\lst{F})} \log [ 1 - \module{ntp}^{\kb}_\params(\tilde{\lst{F}}, d) ]
\end{aligned}
\end{equation}
\glspl{NTP} can also learn \emph{interpretable rules}.
\citet{rocktaschel2017end} show that it is possible to learn rules from data by specifying \emph{rule templates}, such as $\lst{H} \lif \lst{B}$ with $\lst{H} = \xs{\params_{p:}, \var{X}, \var{Y}}$ and $\lst{B} = \xs{\xs{\params_{q:}, \var{X}, \var{Z}}, \xs{\params_{r:}, \var{Z},\var{Y}}}$.
Parameters $\params_{p:}, \params_{q:}, \params_{r:} \in \sR^{k}$, denoting rule-predicate embeddings, can be learned from data by minimising the loss in \cref{eq:loss}, and decoded by searching the closest representation of known predicates.
\section{Efficient Differentiable Reasoning \\ on Large-Scale KBs} \label{sec:scale}
\glspl{NTP} are capable of deductive reasoning, and the proof paths with the highest score can provide human-readable explanations for a given prediction.
However, enumerating and scoring all bounded-depth proof paths for a given goal, as given in \cref{eq:max}, is computationally intractable.
For each goal and sub-goal $\lst{G}$, this process requires to unify $G$ with the representations of \emph{all} rule heads and facts in the \gls{KB}, which quickly becomes computationally prohibitive even for moderately sized \glspl{KB}.
Furthermore, the expansion of a rule like $\rel{p}(\var{X}, \var{Y}) \lif \rel{q}(\var{X}, \var{Z}), \rel{r}(\var{Z}, \var{Y})$ via backward chaining causes an increase of the sub-goals to prove, both because all atoms in the body need to be proven, and because $\var{Z}$ is a free variable with many possible bindings~\citep{rocktaschel2017end}.
We consider two problems -- given a sub-goal $\lst{G}$ such as $\xs{\rel{p}, \const{a}, \const{b}}$, we need to efficiently select
\begin{inparaenum}[\itshape i)\upshape]
\item the $k_f$ \emph{facts} that are most likely to prove a sub-goal $\lst{G}$, and
\item the $k_r$ \emph{rules} to expand to reach a high-scoring proof state.
\end{inparaenum}
\paragraph{Fact Selection.}
Unifying a sub-goal $\lst{G}$ with all facts in the \gls{KB} $\kb$ may not be feasible in practice.
The number of facts in a real-world \gls{KB} can be in the order of millions or billions. For instance, Freebase contains over $637 \times 10^{6}$ facts, while the Google Knowledge Graph contains more than $18 \times 10^{9}$ facts~\citep{DBLP:journals/pieee/Nickel0TG16}.
Identifying the facts $\lst{F} \in \kb$ that yield the maximum proof score for a sub-goal $\lst{G}$ reduces to solving the following optimisation problem:
\begin{equation} \label{eq:problem}
\begin{aligned}
& \module{ntp}^{\kb}_\params(\lst{G}, 1) = \max_{\lst{F} \lif \xs{} \in \kb} \state^{\lst{F}}_\success = \state^{\star}_\success \\
& \text{with} \quad \state^{\lst{F}} = \module{unify}_\params(\lst{F}, \lst{G}, (\emptyset, 1)) \\
\end{aligned}
\end{equation}
Hence, the fact $\lst{F} \in \kb$ that yields the maximum proof score for a sub-goal $\lst{G}$ is the fact $\lst{F}$ that yields the maximum unification score with $\lst{G}$.
Recall that the unification score between a fact $\lst{F}$ and a goal $\lst{G}$ is given by the similarity of their embedding representations $\params_{\lst{F}:}$ and $\params_{\lst{G}:}$, computed via a Gaussian kernel $\krnl(\params_{\lst{F}}, \params_{\lst{G}})$.
Given a goal $\lst{G}$, \glspl{NTP} will compute the unification score between $\lst{G}$ and every fact $\lst{F} \in \kb$ in the \gls{KB}.
This is problematic, since computing the similarity between the representations of the goal $\lst{G}$ and every fact $\lst{F} \in \kb$ is computationally prohibitive -- the number of comparisons is $\gO(|\kb| n)$, where $n$ is the number of (sub-)goals in the proving process.
However, $\module{ntp}^{\kb}_\params(\lst{G}, d)$ only returns the single largest proof score.
This means that, at inference time, we only need the largest proof score for returning the correct output.
Similarly, during training, the gradient of the proof score with respect to the parameters $\params$ can also be calculated exactly by using the single largest proof score:
\begin{equation*}
\begin{aligned}
& \frac{\partial \module{ntp}^{\kb}_\params(\lst{G}, 1)_\success}{\partial \params} = \frac{\partial \max_{\lst{F} \in \kb} \state^{\lst{F}}_\success }{\partial \params} = \frac{\partial \state^{\star}_\success}{\partial \params} \\
& \text{with} \quad \state^{\star}_\success = \max_{\lst{F} \in \kb} \state^{\lst{F}}_\success \\
\end{aligned}
\end{equation*}
In this paper, we propose to efficiently compute $\state^{\star}$, the highest unification score between a given sub-goal $\lst{G}$ and a fact $\lst{F} \in \kb$, by casting it as a \gls{NNS} problem.
This is feasible since the Gaussian kernel used by \glspl{NTP} is a monotonic transformation of the negative Euclidean distance.
Identifying $\state^{\star}$ permits to reduce the number of neural network sub-structures needed for the comparisons between each sub-goal and facts from $\gO(|\kb|)$ to $\gO(1)$.
We use the exact and approximate \gls{NNS} framework proposed by \citet{JDH17} for efficiently searching $\kb$ for the best supporting facts for a given sub-goal.
Specifically we use the exact L2-nearest neighbour search and, for the sake of efficiency, we update the search index every 10 batches, assuming that the small updates made by stochastic gradient descent do not necessarily invalidate previous search indexes.
\paragraph{Rule Selection.}
We use a similar idea for selecting which rules to activate for proving a given goal $\lst{G}$.
We empirically notice that unifying $\lst{G}$ with the closest rule heads, such as $\lst{G} = \xs{\rel{locatedIn}, \const{london}, \const{uk}}$ and $\lst{H} = \xs{\rel{situatedIn}, \var{X}, \var{Y}}$, is more likely to generate high-scoring proof states.
This is a trade-off between symbolic reasoning, where proof paths are expanded only when the heads exactly match with the goals, and differentiable reasoning, where all proof paths are explored.
This prompted us to implement a heuristic that dynamically selects rules among rules sharing the same template during both inference and learning.
In our experiments, this heuristic for selecting proof paths was able to recover valid proofs for a goal when they exist, while drastically reducing the computational complexity of the differentiable proving process.
More formally, we generate a partitioning $\pkb \in 2^\kb$ of the \gls{KB} $\kb$, where each element in $\pkb$ groups all facts and rules in $\kb$ sharing the same template, or high-level structure -- \eg{} an element of $\pkb$ contains all rules with structure $\params_{p:}(\var{X}, \var{Y}) \lif \params_{q:}(\var{X}, \var{Z}), \params_{r:}(\var{Z}, \var{Y})$, with $\params_{p:}, \params_{q:}, \params_{r:} \in \Real^{k}$.~\footnote{Grouping rules with the same structure together makes allows parallel inference to be implemented very efficiently on GPU. This optimisation is also present in \citet{rocktaschel2017end}.}
We then redefine the $\module{or}$ operator as follows:
\begin{equation*} \label{eq:new_or}
\begin{aligned}
& \module{or}^{\kb}_\params( \lst{G}, d, \state) = [ \state' \mid \lst{H} \lif \lst{B} \in \gN_{\gP}(\lst{G}), \gP \in \pkb, \\
& \qquad \qquad \qquad \state' \in \module{and}^{\kb}_\params(\lst{B}, d, \module{unify}_\params(\lst{H}, \lst{G}, \state)) ]
\end{aligned}
\end{equation*}
\noindent where, instead of unifying a sub-goal $\lst{G}$ with all rule heads, we constrain the unification to only the rules where heads are in the neighbourhood $\gN_{\gP}(\lst{G})$ of $\lst{G}$.
\paragraph{Learning to Attend Over Predicates.}
Although \glspl{NTP} can be used for \emph{learning interpretable rules} from data, the solution proposed by \citet{rocktaschel2017end} can be quite inefficient, as the number of parameters associated to rules can be quite large.
For instance, the rule $\lst{H} \lif \lst{B}$, with $\lst{H} = \xs{\params_{p:}, \var{X}, \var{Y}}$ and $\lst{B} = \xs{\xs{\params_{q:}, \var{X}, \var{Z}}, \xs{\params_{r:}, \var{Z},\var{Y}}}$, where $\params_{p:}, \params_{q:}, \params_{r:} \in \sR^{k}$, introduces $3 k$ parameters in the model, where $k$ denotes the embedding size, and it may be computationally inefficient to learn each of the embedding vectors if $k$ is large.
We propose using an \emph{attention mechanism}~\citep{bahdanau2015neural} for attending over known predicates for defining the rule-predicate embeddings $\params_{p:}, \params_{q:}, \params_{r:}$.
Let $\set{R}$ be the set of known predicates, and let $\emR \in \sR^{|\set{R}| \times k}$ be a matrix representing the embeddings for the predicates in $\set{R}$.
We define $\params_{p:}$ as $\params_{p:} = \softmax(\rva_{p:})^{\top} \emR$.
\noindent where $\rva_{p:} \in \sR^{|\set{R}|}$ is a set of trainable \emph{attention weights} associated with the predicate $p$.
This sensibly improves the parameter efficiency of the model in cases where the number of known predicates is low, \ie{} $|\set{R}| \ll k$, by introducing $c |\set{R}|$ parameters for each rule rather than $c k$, where $c$ is the number of trainable predicate embeddings in the rule.
\section{Jointly Reasoning on Knowledge Bases \\ and Natural Language} \label{sec:text}
In this section, we show how \glspl{NaNTP} can jointly reason over \glspl{KB} and natural language corpora.
In the following, we assume that our \gls{KB} $\kb$ is composed of facts, rules, and \emph{textual mentions}.
A fact is composed of a predicate symbol and a sequence of arguments, \eg \ $\xs{\rel{locationOf}, \const{London}, \const{UK}}$.
On the other hand, a \emph{mention} is a textual pattern between two co-occurring entities in the \gls{KB}~\citep{DBLP:conf/emnlp/ToutanovaCPPCG15}, such as ``\const{London} \emph{is located in the} \const{UK}''.
We represent mentions jointly with facts and rules in $\kb$ by considering each textual surface pattern linking two entities as a new predicate, and embedding it in a $d$-dimensional space by means of an end-to-end differentiable reading component.
For instance, the sentence ``United Kingdom borders with Ireland'' can be translated into the following mention: $\xs{\xs{\rel{[arg1]}, \rel{borders}, \rel{with}, \rel{[arg2]}}, \const{UK}, \const{ireland}}$, by first identifying sentences or paragraphs containing \gls{KB} entities, and then considering the textual surface pattern connecting such entities as an extra relation type.
While predicates in $\set{R}$ are encoded by a look-up operation to a predicate embedding matrix $\emR \in \sR^{|\set{R}| \times k}$, textual surface patterns are encoded by an $\module{encode}_{\params} : \datadom{\set{V}^{*}} \to \tensordom{\sR^{k}}$ module,  where $\set{V}$ is the vocabulary of words and symbols occurring in textual surface patterns.
More formally, given a textual surface pattern $t \in \set{V}^{*}$ -- such as $t = \xs{\rel{[arg1]}, \rel{borders}, \rel{with}, \rel{[arg2]}}$ -- the $\module{encode}_{\params}$ module first encodes each token $w$ in $t$ by means of a token embedding matrix $\emV \in \sR^{|\set{V}| \times k'}$, resulting in a pattern matrix $\emW_{t} \in \sR^{|t| \times k'}$.
Then, the module produces a textual surface pattern embedding vector $\params_{t:} \in \sR^{k}$ from $\emW_{t}$ by means of an end-to-end differentiable encoder.
For assessing whether a simple encoder architecture can already provide benefits to the model, we use an $\module{encode}_{\params}$ module that aggregates the embeddings of the tokens composing a textual surface pattern via mean pooling: $\module{encode}_{\params}(t) = \frac{1}{|t|}\sum_{w \in t} \emV_{w\cdot} \in \sR^{k}$.
Albeit the encoder 
can be implemented by using other differentiable architectures, for this work we opted for a simple but still very effective Bag of Embeddings model~\citep{white2015well,arora2016simple} showing that, even in this case, the model achieves very accurate results.
\section{Related Work} \label{sec:related}
\begin{figure}[t]
\centering
\includegraphics[width=\columnwidth]{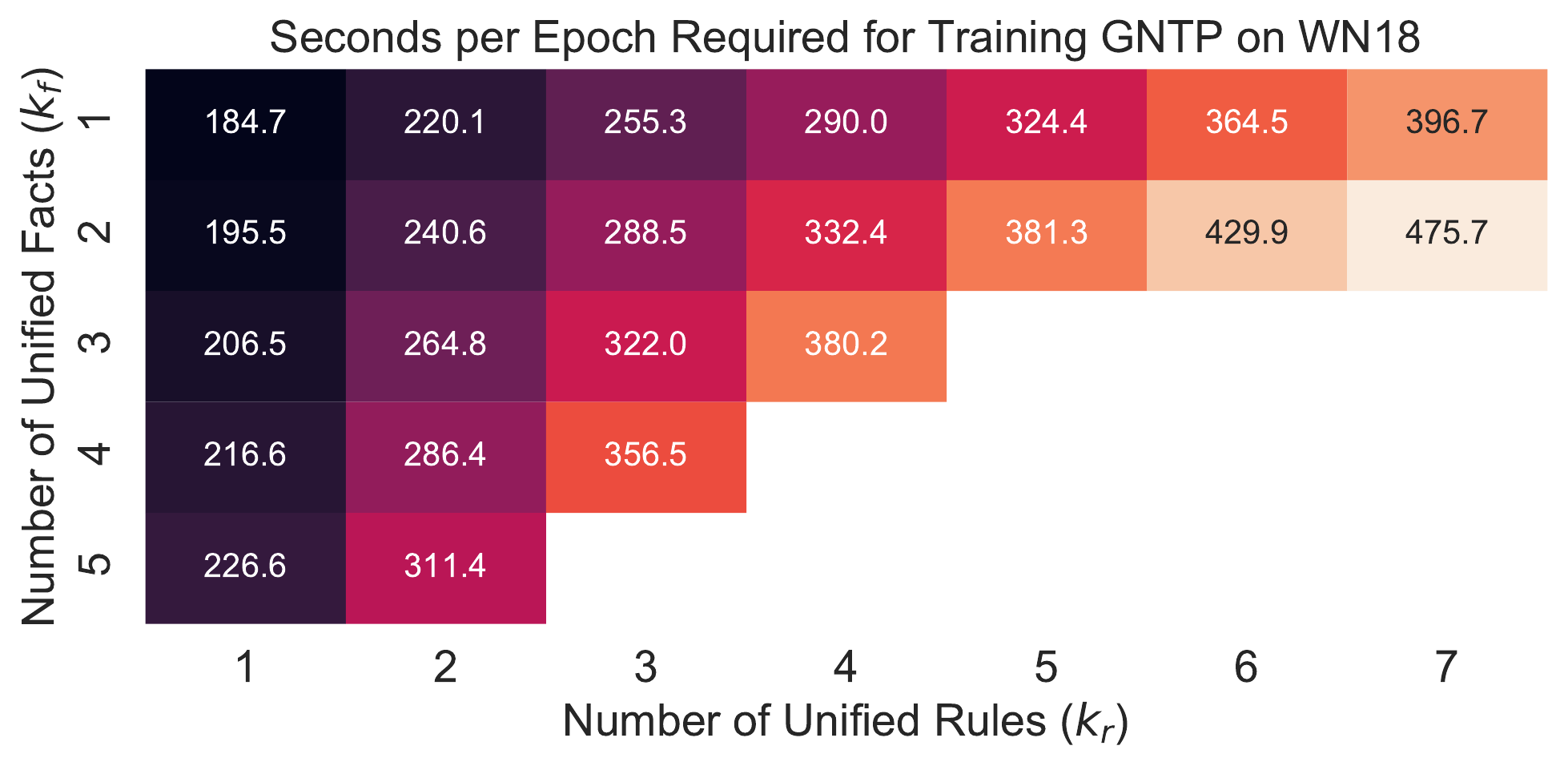}
\caption{Number of seconds per epoch required for training on the WN18 dataset using batches of 1000 examples on a GPU. Missing entries denote out-of-memory errors.}
\label{fig:scalability}
\end{figure}
A notable corpus of literature aims at addressing the limitations of neural architectures in terms of generalisation and reasoning abilities.
A line of research consists of enriching neural network architectures with a differentiable \emph{external memory}~\citep{DBLP:conf/nips/SukhbaatarSWF15,DBLP:journals/corr/GravesWD14,joulin2015inferring,grefenstette2015learning,DBLP:journals/corr/KaiserS15}.
The underlying idea is that a neural network can learn to represent and manipulate complex data structures, thus disentangling the algorithmic part of the process from the representation of the inputs.
By doing so, it becomes possible to train such models from enriched supervision signals, such as from \emph{program traces} rather than simple input-output pairs.
A related field is \emph{differentiable interpreters}---program interpreters where declarative or procedural knowledge is compiled into a neural network architecture~\citep{DBLP:conf/icml/BosnjakRNR17,rocktaschel2017end,evans2018learning}.
This family of models allows imposing strong inductive biases on the models by partially defining the program structure used for constructing the network, \eg{}, in terms of instruction sets or rules.
A major drawback of differentiable interpreters, however, is their computational complexity, so far deeming them unusable except for smaller learning problems.
\citet{rae2016scaling} use an approximate nearest neighbour data structures for sparsifying read operations in memory networks.
\citet{DBLP:conf/naacl/RiedelYMM13} pioneered the idea of jointly embedding \gls{KB} facts and textual mentions in shared embedding space, by considering mentions as additional relations in a \gls{KB} factorisation setting, and more elaborate mention encoders were investigated by \citet{DBLP:conf/eacl/McCallumNV17}.
Our work is also related to path encoding models~\citep{DBLP:conf/eacl/McCallumNDB17} and random walk approaches~\citep{lao2011random,gardner2014incorporating}, both of which lack a rule induction mechanisms, and to approaches combining observable and latent features of the graph~\citep{DBLP:conf/nips/NickelJT14,DBLP:conf/sac/MinervinidFE16}.
Lastly, our work is related to \citet{DBLP:conf/nips/YangYC17}, a scalable rule induction approach for \gls{KB} completion, but has not been applied to textual surface patterns.
\begin{figure*}[t]
\centering
\begin{subfigure}{0.5\linewidth}
 \centering
 \includegraphics[width=\linewidth]{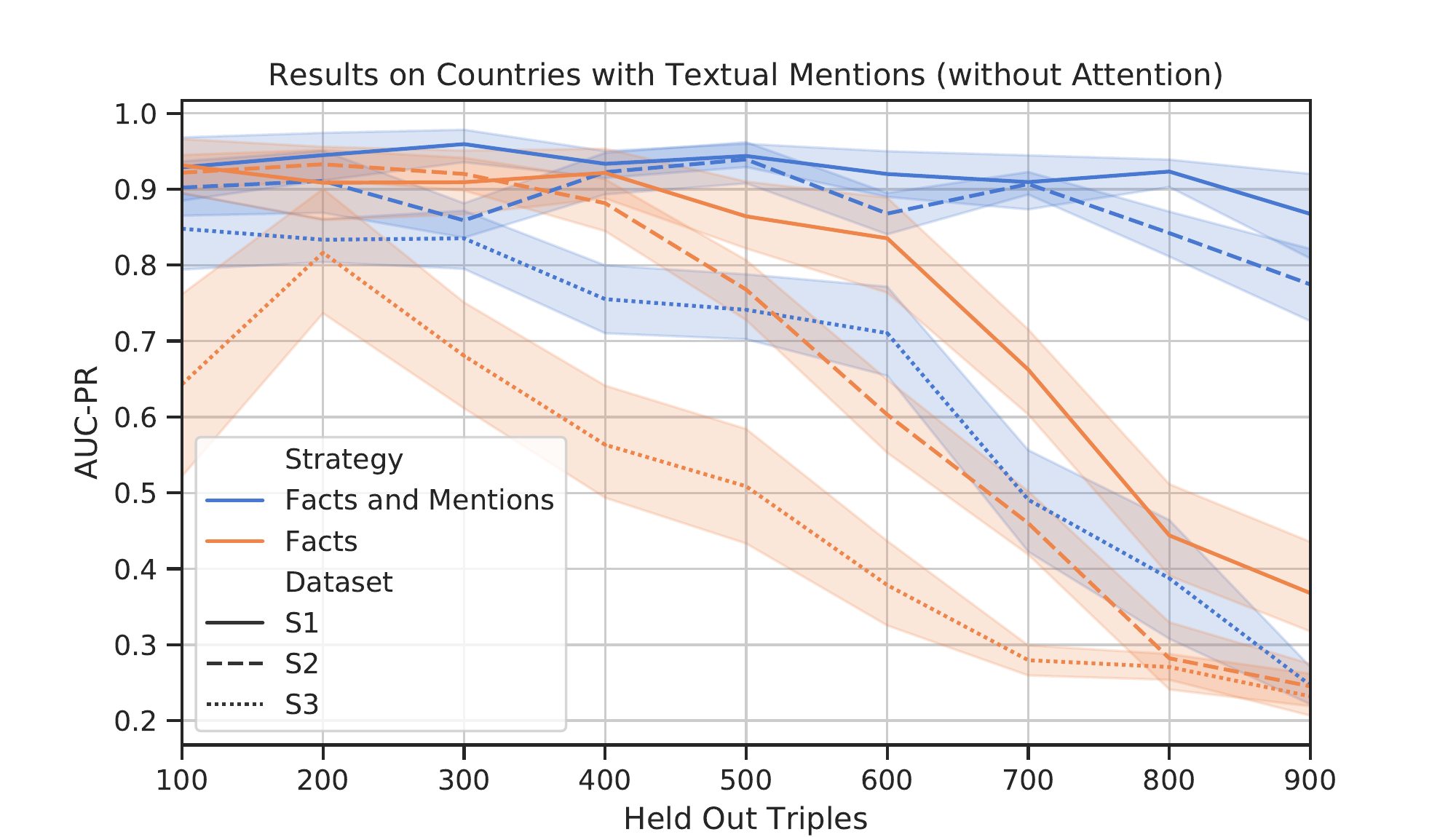}
\end{subfigure}%
\begin{subfigure}{0.5\linewidth}
 \centering
 \includegraphics[width=\linewidth]{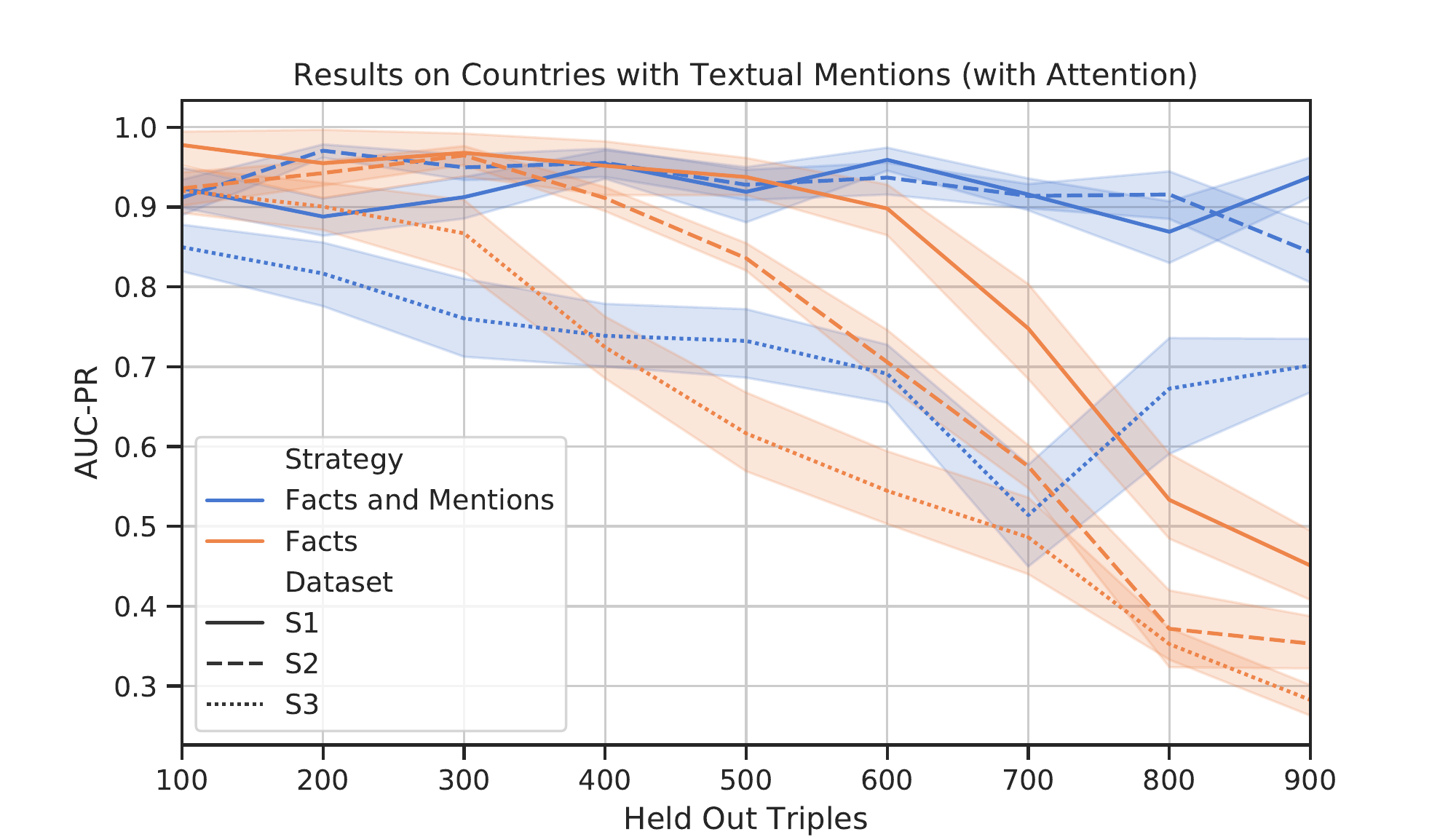}
\end{subfigure}
\caption{
\glspl{NaNTP} on Countries with generated mentions. We replaced a varying number of relations with textual mentions and integrated them by encoding the mentions using a text encoder (\emph{Facts and Mentions}) and by simply adding them to the \gls{KB} (\emph{Facts}).
Two figures contrast the effects of rule learning without attention (left) and with it (right).
}
\label{fig:countries}
\end{figure*}
\section{Experiments} \label{sec:experiments}

\paragraph{Datasets and Evaluation Protocols.}
We report the results of experiments on benchmark datasets --- Countries~\citep{bouchard2015approximate}, Nations, UMLS, and Kinship~\citep{kemp2006learning} --- following the same evaluation protocols as \citet{rocktaschel2017end}.
Furthermore, since \glspl{NaNTP} allows to experiment on significantly larger datasets, we also report results on the WN18~\citep{DBLP:conf/nips/BordesUGWY13}, WN18RR~\citep{DBLP:conf/aaai/DettmersMS018} and FB122~\citep{DBLP:conf/emnlp/GuoWWWG16} datasets. 
Results are reported in terms of the \gls{AUC-PR}~\citep{DBLP:conf/icml/DavisG06}, \gls{MRR}, and HITS@$m$~\citep{DBLP:conf/nips/BordesUGWY13}.
Datasets and hyperparameters are described in the Appendix.~\footnote{\label{appendix}The Appendix can be found at \url{https://github.com/uclnlp/gntp}}
\paragraph{Baselines.}
On benchmark datasets, we compare \glspl{NaNTP} with \glspl{NTP} and two other neuro-symbolic reasoning systems, MINERVA~\citep{DBLP:journals/corr/abs-1711-05851}, which employs a reinforcement learning algorithm to reach answers by traversing the \gls{KB} graph, and NeuralLP~\citep{DBLP:conf/nips/YangYC17}, which compiles inference tasks in a sequence of differentiable operations.
In addition, we consider DistMult~\citep{yang15:embedding} and ComplEx~\citep{DBLP:conf/icml/TrouillonWRGB16}, two state-of-the-art black-box neural link predictors suited for large datasets.
\paragraph{Run-Time Evaluation.}
To assess the benefits of \glspl{NaNTP} in terms of computational complexity and range of applications, we consider the best hyperparameters we found for the WN18 dataset, and measured the time needed for each training epoch varying the number of unified facts and rules during inference. 
Results, outlined in \cref{fig:scalability}, show that learning on WN18 quickly becomes infeasible by increasing the number of unified facts and rules.
\glspl{NTP} are a special case of \glspl{NaNTP} where, during the forward pass, there is no pruning of the proof paths.
From \cref{fig:scalability} we can see that even for \glspl{KB} a fraction the size of WordNet and Freebase, \glspl{NTP} rapidly run out of memory, deeming them inapplicable to reasonably sized \glspl{KB}.
Instead, sensible pruning of proof paths in \glspl{NaNTP} drastically increases the efficiency of both the learning and the inference process, allowing to train on large \glspl{KB} like WordNet.
We refer to the Appendix\footnotemark[\getrefnumber{appendix}] for additional experiments showing run-time improvements by several orders of magnitude.
\begin{table*}[t]
\caption{
Comparison of \glspl{NaNTP}, \glspl{NTP}, NeuralLP~\citep{DBLP:conf/nips/YangYC17}, and MINERVA~\citep{DBLP:journals/corr/abs-1711-05851} (from \citet{DBLP:journals/corr/abs-1711-05851}) on benchmark datasets, with and without attention.
}
\label{tab:benchmark}
\centering
\resizebox{\textwidth}{!}{
\begin{tabular}{cclcccccl}
\toprule
\multicolumn{2}{c}{\multirow{4}{*}{\bf Datasets}} & \multirow{4}{*}{\bf Metrics} & \multicolumn{5}{c}{\bf Models} & \multirow{4}{*}{\bf \qquad \qquad Rules Learned by \gls{NaNTP}} \\
\cmidrule{4-8}
& & & \multirow{2}{*}{\bf \gls{NTP}~\footnotemark[7]} & \multicolumn{2}{c}{\bf \gls{NaNTP}} & \multirow{2}{*}{\bf NeuralLP} & \multirow{2}{*}{\bf MINERVA} \\
\cmidrule{5-6}
& & & & {\bf Standard} & {\bf Attention} \\
\midrule
\multirow{3}{*}{\bf Countries} & S1 & \multirow{3}{*}{AUC-PR} & 90.83 $\pm$ 15.4 & 99.98 $\pm$ 0.05 & {\bf 100.0 $\pm$ 0.0} & {\bf 100.0 $\pm$ 0.0} & {\bf 100.0 $\pm$ 0.0} & \rel{locatedIn}(\var{X},\var{Y}) \lif \rel{locatedIn}(\var{X},\var{Z}), \rel{locatedIn}(\var{Z},\var{Y}) \\
& S2 & & 87.40 $\pm$ 11.7 & 90.82 $\pm$ 0.88 & \textbf{93.48 $\pm$ 3.29} & 75.1 $\pm$ 0.3 & 92.36 $\pm$ 2.41 & \rel{neighborOf}(\var{X},\var{Y}) \lif \rel{neighborOf}(\var{X},\var{Z}), \rel{locatedIn}(\var{Z},\var{Y}) \\
& S3 & & 56.68 $\pm$ 17.6 & 87.70 $\pm$ 4.79 & 91.27 $\pm$ 4.02 & 92.20 $\pm$ 0.2 & {\bf 95.10 $\pm$ 1.20} & \rel{neighborOf}(\var{X},\var{Y}) \lif \rel{neighborOf}(\var{Y},\var{X}) \\
\midrule
\multirow{4}{*}{\bf Kinship}
& & MRR & 0.35 & 0.719 & {\bf 0.759} & 0.619 & 0.720 & \rel{term0}(\var{X}, \var{Y}) \lif \rel{term0}(\var{Y}, \var{X}) \\
& & HITS@1 & 0.24 & 0.586 & {\bf 0.642} & 0.475 & 0.605 & \rel{term4}(\var{X}, \var{Y}) \lif \rel{term4}(\var{Y}, \var{X}) \\
& & HITS@3 & 0.37 & 0.815 & {\bf 0.850} & 0.707 & 0.812 & \rel{term13}(\var{X},\var{Y}) \lif \rel{term13}(\var{X}, \var{Z}), \rel{term10}(\var{Z}, \var{Y}) \\
& & HITS@10 & 0.57  & 0.958 & {\bf 0.959} & 0.912 & 0.924 & \rel{term2}(\var{X},\var{Y}) \lif \rel{term4}(\var{X}, \var{Z}), \rel{term7}(\var{Z}, \var{Y}) \\
\midrule
\multirow{4}{*}{\bf Nations}
& & MRR & 0.61 & {\bf 0.658} & 0.645 & --- & --- & \rel{commonbloc1}(\var{X}, \var{Y}) \lif \rel{relngo}(\var{Y}, \var{X}) \\
& & HITS@1 & 0.45 & {\bf 0.493} & 0.490 & --- & --- & \rel{timesincewar}(\var{X},\var{Y}) \lif \rel{independence}(\var{X},\var{Y}) \\
& & HITS@3 & 0.73 & {\bf 0.781} & 0.736 & --- & --- &  \rel{unweightedunvote}(\var{X},\var{Y}) \lif \rel{relngo}(\var{X},\var{Y}) \\
& & HITS@10 & 0.87 & {\bf 0.985} & 0.975 & --- & --- & \rel{ngo}(\var{X}, \var{Y}) \lif \rel{independence}(\var{Y}, \var{X}) \\
\midrule
\multirow{4}{*}{\bf UMLS}
& & MRR & 0.80 & 0.841 & {\bf 0.857} & 0.778 & 0.825 & \rel{isa}(\var{X},\var{Y}) \lif \rel{isa}(\var{X},\var{Z}), \rel{isa}(\var{Z},\var{Y}) \\
& & HITS@1 & 0.70 & 0.732 & {\bf 0.761} & 0.643 & 0.728 & \rel{complicates}(\var{X},\var{Y}) \lif \rel{affects}(\var{X},\var{Y}) \\
& & HITS@3 & 0.88 & {\bf 0.941} & {\bf 0.947} & 0.869 & 0.900 & \rel{affects}(\var{X}, \var{Y}) \lif \rel{affects}(\var{X}, \var{Z}), \rel{affects}(\var{Z}, \var{Y}) \\
& & HITS@10 & 0.95 & {\bf 0.986} & 0.983 & 0.962 & 0.968 & 
\rel{process\_of}(\var{X},\var{Y}) \lif \rel{affects}(\var{X},\var{Y}) \\
\bottomrule
\end{tabular}
}
\end{table*}
\paragraph{Link Prediction Results.} \label{sec:benchmark}

We compare \glspl{NaNTP} and \glspl{NTP} on a set of link prediction benchmarks, also used in \citet{rocktaschel2017end}.
Results, presented in \cref{tab:benchmark}, show that \glspl{NaNTP} achieves better or on-par results in comparison with \glspl{NTP} and baselines MINERVA~\citep{DBLP:journals/corr/abs-1711-05851} and NeuralLP~\citep{DBLP:journals/corr/abs-1711-05851}, consistently through all benchmark datasets. 
We can also see that models learned by \glspl{NaNTP} are \emph{interpretable}: in \cref{tab:benchmark} we show the decoded rules learned by the model, and learn about the domain at hand.
For instance, we can see that on UMLS, a biomedical \gls{KB}, the \rel{isa} and \rel{affects} relation are transitive.
\paragraph{Experiments with Generated Mentions.}
For evaluating different strategies of integrating textual surface patterns, in the form of mentions, in \glspl{NTP}, we proceeded as follows.
We replaced a varying number of training set triples from each of the Countries S1-S3 datasets with human-generated textual mentions (for more details, see Appendix).\footnotemark[6]
For instance, the fact $\rel{neighbourOf}(\const{UK}, \const{Ireland})$ may be replaced by the textual mention ``\const{UK}\rel{ is neighbouring with }\const{Ireland}''.
The entities $\const{UK}$ and $\const{Ireland}$ become the arguments, while the text between them is treated as a new logic predicate, forming a new fact  $\text{``} \var{X}\ \text{\sffamily is neighbouring with } \var{Y} \text{''} (\const{UK}, \const{Ireland})$.
Then, we evaluate two ways of integrating textual mentions in \glspl{NaNTP}:
\begin{inparaenum}[\itshape i)\upshape]
\item adding them as facts to the \gls{KB}, and
\item parsing the mention by means of an encoder.
\end{inparaenum}
The results, presented in \cref{fig:countries}, show that the proposed encoding module yields consistent improvements of the ranking accuracy in comparison to simply adding the mentions as facts to the \gls{KB}.
This is especially evident in cases where the number of held-out facts is higher, as it is often the case in real-world use cases, where there is an abundance of text but the \glspl{KB} are sparse and incomplete~\citep{DBLP:journals/pieee/Nickel0TG16}.
\glspl{NaNTP} are extremely efficient at learning rules involving both \emph{logic atoms and textual mentions}.
For instance, by analysing the learned models and their explanations, we can see that \glspl{NaNTP} learn rules such as
{\renewcommand{\arraystretch}{1.9}
\begin{center}
\resizebox{0.47\textwidth}{!}{
\begin{tabular}{{@{}l}}
$\rel{neighborOf}(\var{X}, \var{Y}) \lif \text{``} \var{Y}\ \text{\sffamily is a neighboring state to } \var{X} \text{''} (\var{X}, \var{Y})$ \\
$\begin{aligned}
\rel{locatedIn}(\var{X}, \var{Y}) \lif &\text{``} \var{X}\ \text{\sffamily is a neighboring state to } \var{Z} \text{''} (\var{X}, \var{Z}), \\&\text{ ``} \var{Z}\ \text{\sffamily is located in } \var{Y} \text{''} (\var{Z}, \var{Y})
\end{aligned}$ \\
\end{tabular}
}
\end{center}
}
\noindent and leverage them during their reasoning process, providing human-readable explanations for a given prediction.
\footnotetext[7]{Results reported in \citet{rocktaschel2017end} were calculated with an incorrect evaluation function, causing artificially better results. We corrected the issues, and recalculated the results.}
\begin{table*}
\centering
\caption{
Link prediction results on the Test-I, Test-II and Test-ALL on FB122.
Note that KALE, \emph{ASR} methods, and \textsc{KBlr} have access to a set of rules provided by \citet{DBLP:conf/emnlp/GuoWWWG16}, while neural link predictors and \glspl{NaNTP} do not.
Test-II (6,186 triples) denotes a subset of FB122 that can be inferred via logic rules, while Test-I (5,057 triples) denotes all other test triples.
We can see that, even without providing any rule to the model, \glspl{NaNTP} yields better ranking results in comparison with neural link prediction models---since it is able to learn such rules from data---and it is comparable with models that can leverage the provided rules.
} \label{tab:fb122}
\resizebox{\textwidth}{!}{
\begin{tabular}{clcccccccccccccc}
\toprule
& & \multicolumn{4}{c}{Test-I} & & \multicolumn{4}{c}{Test-II} & & \multicolumn{4}{c}{Test-ALL}\\
\cline{3-6}\cline{8-11}\cline{13-15}
& & \multicolumn{3}{c}{Hits@N (\%)} & \multirow{2}{*}{MRR} & & \multicolumn{3}{c}{Hits@N (\%)} & \multirow{2}{*}{MRR} & & \multicolumn{3}{c}{Hits@N (\%)} & \multirow{2}{*}{MRR}
\\
\cline{3-5}\cline{8-10}\cline{13-15}
& & 3 & 5 & 10 & & \ & 3 & 5 & 10 & \ & & 3 & 5 & 10 & \\
\midrule
\multirow{5}{*}{\rotatebox[origin=c]{90}{\parbox[c]{1.5cm}{\centering {\bf With Rules}}}}
& KALE-Pre~\citep{DBLP:conf/emnlp/GuoWWWG16} & 35.8 & 41.9 & 49.8 & 0.291 & & 82.9 & 86.1 & 89.9 & 0.713 & & 61.7 & 66.2 & 71.8 & 0.523 \\
& KALE-Joint~\citep{DBLP:conf/emnlp/GuoWWWG16} & {\bf 38.4} & {\bf 44.7} & {\bf 52.2} & 0.325 & & 79.7 & 84.1 & 89.6 & 0.684 & & 61.2 & 66.4 & 72.8 & 0.523 \\
& \emph{ASR}-DistMult~\citep{DBLP:conf/uai/MinerviniDRR17} & 36.3 & 40.3 & 44.9 & 0.330 & & 98.0 & 99.0 & 99.2 & 0.948 & & 70.7 & 73.1 & 75.2 & 0.675 \\
& \emph{ASR}-ComplEx~\citep{DBLP:conf/uai/MinerviniDRR17} & 37.3 & 41.0 & 45.9 & {\bf 0.338} & & {\bf 99.2} & {\bf 99.3} & {\bf 99.4} & {\bf 0.984} & & 71.7 & 73.6 & 75.7 & 0.698 \\
& \textsc{KBlr}~\citep{DBLP:conf/uai/Garcia-DuranN18} & -- & -- & -- & -- & & -- & -- & -- & -- & & {\bf 74.0} & {\bf 77.0} & {\bf 79.7} & {\bf 0.702} \\
\midrule
\multirow{4}{*}{\rotatebox[origin=c]{90}{\parbox[c]{1.5cm}{\centering {\bf Without Rules}}}}
& TransE~\citep{DBLP:conf/nips/BordesUGWY13} & 36.0 & 41.5 & 48.1 & 0.296 & & 77.5 & 82.8 & 88.4 & 0.630 & & 58.9 & 64.2 & 70.2 & 0.480 \\
& DistMult~\citep{yang15:embedding} & 36.0 & 40.3 & 45.3 & 0.313 & & {\bf 92.3} & {\bf 93.8} & {\bf 94.7} & 0.874 & & {\bf 67.4} & {\bf 70.2} & {\bf 72.9} & 0.628 \\
& ComplEx~\citep{DBLP:conf/icml/TrouillonWRGB16} & {\bf 37.0} & {\bf 41.3} & {\bf 46.2} & {\bf 0.329} & & 91.4 & 91.9 & 92.4 & {\bf 0.887} & & 67.3 & 69.5 & 71.9 & {\bf 0.641} \\
& \glspl{NaNTP} & 33.7 & 36.9 & 41.2 & 0.313 & & \marktopleft{c1}{\bf 98.2} & {\bf 99.0} & {\bf 99.3} & {\bf 0.977}\markbottomright{c1} & & \marktopleft{c2}{\bf 69.2} & {\bf 71.1} & {\bf 73.2} & {\bf 0.678}\markbottomright{c2} \\
\bottomrule
\end{tabular}
}
\end{table*}
\begin{tikzpicture}[remember picture,overlay]
\foreach \Val in {up,down}
{
}
\end{tikzpicture}

\begin{table*}[t]
\caption{Explanations, in terms of rules and supporting facts, for the queries in the validation set of WN18 provided by \glspl{NaNTP} by looking at the proof paths yielding the largest proof scores.}
\label{tab:explanations}
    \centering
	\resizebox{\textwidth}{!}{
	\begin{tabular}{crcl}
	    \toprule
	    
	    \multicolumn{2}{r}{\bf Query} & {\bf Score} $\state_\success$ & {\bf Proofs / Explanations} \\
	    
	    \midrule

\multirow{8}{*}{\begin{turn}{90}{\bf{WN18}}\end{turn}}
& \multirow{3}{*}{$\rel{part\_of}(\const{congo.n.03}, \const{africa.n.01})$} &  0.995 &  {$\!\begin{aligned} 
\rel{part\_of}(\var{X}, \var{Y}) \lif &\rel{has\_part}(\var{Y}, \var{X}) \qquad \rel{has\_part}(\const{africa.n.01}, \const{congo.n.03}) \end{aligned}$}\\

\arrayrulecolor{black!30}\cmidrule{3-4}

& &  0.787 & 
{$\!\begin{aligned}
\rel{part\_of}(\var{X}, \var{Y}) \lif &\rel{instance\_hyponym}(\var{Y}, \var{X}) \\
&\rel{instance\_hyponym}(\const{african\_country.n.01}, \const{congo.n.03})
 \end{aligned}$}\\

\arrayrulecolor{black}\cmidrule{2-4}

& \multirow{1}{*}{$\rel{hyponym}(\const{extinguish.v.04}, \const{decouple.v.03})$} &  0.987 & {$\!\begin{aligned}
\rel{hyponym}(\var{X}, \var{Y}) \lif &\rel{hypernym}(\var{Y}, \var{X}) \qquad \rel{hypernym}(\const{decouple.v.03}, \const{extinguish.v.04})
 \end{aligned}$}\\

\arrayrulecolor{black}\cmidrule{2-4}

& $\rel{has\_part}(\const{texas.n.01}, \const{odessa.n.02})$ & 0.961 & {$\!\begin{aligned}
\rel{has\_part}(\var{X}, \var{Y}) \lif &\rel{part\_of}(\var{Y}, \var{X}) \qquad \rel{part\_of}(\const{odessa.n.02}, \const{texas.n.01})
 \end{aligned}$}\\

		\bottomrule
	\end{tabular}
	}
\end{table*}
\subsection{Results on Freebase and WordNet} \label{ssec:explanations}
Link prediction results for FB122 are summarised in \cref{tab:fb122}.
The FB122 dataset proposed by \citet{DBLP:conf/emnlp/GuoWWWG16} is fairly large scale: it comprises 91,638 triples, 9,738 entities, and 122 relations, as well as 47 rules that can be leveraged by models for link prediction tasks.
For such a reason, we consider a series of models that can leverage the presence of such rules, namely KALE~\citep{DBLP:conf/emnlp/GuoWWWG16}, DistMult and ComplEx using Adversarial Sets (\emph{ASR})~\citep{DBLP:conf/uai/MinerviniDRR17}---a method for incorporating rules in neural link predictors via adversarial training---and the recently proposed \textsc{KBlr}~\citep{DBLP:conf/uai/Garcia-DuranN18}.
Note that, unlike these methods, \glspl{NaNTP} do not have access to such rules and need to learn them from data.
\cref{tab:fb122} shows that \gls{NaNTP}, whilst not having access to rules, performs significantly better than neural link predictors, and on-par with methods that have access to all rules.
In particular, we can see that on Test-II, a subset of FB122 directly related to logic rules, \gls{NaNTP} yields competitive results.
\gls{NaNTP} is able to induce rules relevant for accurate predictions, such as:

{\renewcommand{\arraystretch}{1.2}
\begin{center}
\resizebox{0.47\textwidth}{!}{
\begin{tabular}{{@{}l}}
\rel{timeZone}(\var{X}, \var{Y}) \lif \rel{containedBy}(\var{X}, \var{Z}), \rel{timeZone}(\var{Z}, \var{Y}). \\
\rel{nearbyAirports}(\var{X}, \var{Y}) \lif \rel{containedBy}(\var{X}, \var{Z}), \rel{contains}(\var{Z}, \var{Y}). \\
\rel{children}(\var{X}, \var{Y}) \lif \rel{parents}(\var{Y}, \var{X}). \\
\rel{spouse}(\var{X}, \var{Y}) \lif \rel{spouse}(\var{Y}, \var{X}). \\
\end{tabular}
}
\end{center}
}

We also evaluate \gls{NaNTP} on WN18~\citep{DBLP:conf/nips/BordesUGWY13} and WN18RR~\citep{DBLP:conf/aaai/DettmersMS018}.
In terms of ranking accuracy, \glspl{NaNTP} is comparable to state-of-the-art models, such as ComplEx and \textsc{KBlr}.
In \citet{DBLP:conf/uai/Garcia-DuranN18} authors report a $94.2$ MRR for ComplEx and $93.6$ MRR for \textsc{KBlr}, while NeuralLP~\citep{DBLP:conf/nips/YangYC17} achieves $94.0$, with hits@10 equal to $94.5$.
\gls{NaNTP} achieves $94.2$ MRR and $94.31$, $94.41$, $94.51$ hits@3, 5, 10, which is on par with state-of-the-art neural link prediction models, while being interpretable via proof paths. \cref{tab:explanations} shows an excerpt of validation triples together with their \gls{NaNTP} proof scores and associated proof paths for WN18.
On WN18RR, \gls{NaNTP} with MRR of $43.4$ performs close to ComplEx~\citep{DBLP:conf/aaai/DettmersMS018} ($44.0 $ MRR) but lags behind NeuralLP ($46.3$ MRR).

We can see that \glspl{NaNTP} is capable of learning and utilising rules, such as $\protect{\rel{has\_part}(\var{X}, \var{Y})} \lif \rel{part\_of}(\var{Y}, \var{X})$, and $\rel{hyponym}(\var{X}, \var{Y}) \lif \rel{hypernym}(\var{Y}, \var{X})$.
Interestingly, \gls{NaNTP} is able to find non-trivial explanations for a given fact, based on the similarity between entity representations.
For instance, it can explain that \const{congo} is part of \const{africa} by leveraging the semantic similarity with \const{african\_country}. 
\section{Conclusions} \label{sec:conclusions}
\glspl{NTP} combine the strengths of rule-based and neural models but, so far, they were unable to reason over large \glspl{KB} and natural language.
In this paper, we overcome such limitations by considering only the subset of proof paths associated with the largest proof scores during the construction of a dynamic computation graph.
The proposed model, \gls{NaNTP}, is more computationally efficient by several orders of magnitude, while achieving similar or better predictive performance than \glspl{NTP}.
\glspl{NaNTP} enable end-to-end differentiable reasoning on large \glspl{KB} and natural language texts, by embedding logic atoms and textual mentions in the same embedding space.
Furthermore, \glspl{NaNTP} are interpretable and can provide explanations in terms of logic proofs at scale.

\bibliography{bibliography}
\bibliographystyle{aaai}

\appendix

\part*{Appendix}

\begin{table*}[t]
\centering
\resizebox{\textwidth}{!}{
\begin{tabular}{cc}
\toprule
{\bf Predicate Name} & {\bf Mentions} \\

\midrule

\multirow{6}{*}{$\rel{locatedIn}(a, b)$} & \multicolumn{1}{p{12cm}}{\raggedright $a$ is located in  $b$, $a$ is situated in  $b$, $a$ is placed in  $b$, $a$ is positioned in  $b$, $a$ is sited in  $b$, $a$ is currently in  $b$, $a$ can be found in  $b$, $a$ is still in  $b$, $a$ is localized in  $b$, $a$ is present in  $b$, $a$ is contained in  $b$, $a$ is found in  $b$, $a$ was located in  $b$, $a$ was situated in  $b$, $a$ was placed in  $b$, $a$ was positioned in  $b$, $a$ was sited in  $b$, $a$ was currently in  $b$, $a$ used to be found in  $b$, $a$ was still in  $b$, $a$ was localized in  $b$, $a$ was present in  $b$, $a$ was contained in  $b$, $a$ was found in  $b$} \\

\midrule

\multirow{4}{*}{$\rel{neighborOf}(a, b)$} &
\multicolumn{1}{p{12cm}}{\raggedright $a$ is adjacent to  $b$, $a$ borders with  $b$, $a$ is butted against  $b$, $a$ neighbours  $b$, $a$ is a neighbor of  $b$, $a$ is a neighboring country of  $b$, $a$ is a neighboring state to  $b$, $a$ was adjacent to  $b$, $a$ borders  $b$, $a$ was butted against  $b$, $a$ neighbours with  $b$, $a$ was a neighbor of  $b$, $a$ was a neighboring country of  $b$, $a$ was a neighboring state to $b$} \\

\bottomrule
\end{tabular}
}
\caption{Mentions used for replacing a varying number of training triples in the Countries S1, S2, and S3 datasets.}
\label{tab:mentions}
\end{table*}

\begin{table*}[t]
\caption{Examples of the clauses used for Freebase (FB122) and WordNet (WN18).} \label{tab:rules}
\centering
\resizebox{\textwidth}{!}{
\begin{tabular}{lr}
\toprule
\multicolumn{2}{l}{$\rel{/people/person/languages}(\var{X},\var{Z}) \lif \rel{/people/person/nationality}(\var{X},\var{Y}), \rel{/location/country/official\_language}(\var{Y},\var{Z})$} \\
\multicolumn{2}{l}{$\rel{/location/contains}(\var{X},\var{Z}) \lif \rel{/country/administrative\_divisions}(\var{X},\var{Y}),  \rel{/administrative\_division/capital}(\var{Y},\var{Z})$} \\
\multicolumn{2}{l}{$\rel{/location/location/contains}(\var{X},\var{Y}) \lif \rel{/location/country/capital}(\var{X},\var{Y})$} \\
\midrule
$\rel{\_hyponym}(\var{Y}, \var{X}) \lif \rel{\_hypernym}(\var{X}, \var{Y})$ & $\rel{\_hypernym}(\var{Y}, \var{X}) \lif \rel{\_hyponym}(\var{X}, \var{Y})$ \\
$\rel{\_part\_of}(\var{Y}, \var{X}) \lif \rel{\_has\_part}(\var{X}, \var{Y})$ & $\rel{\_has\_part}(\var{Y}, \var{X}) \lif \rel{\_part\_of}(\var{X}, \var{Y})$ \\
\bottomrule
\end{tabular}
}
\end{table*}
\begin{figure*}[t!]
\centering
\includegraphics[width=1.0\textwidth]{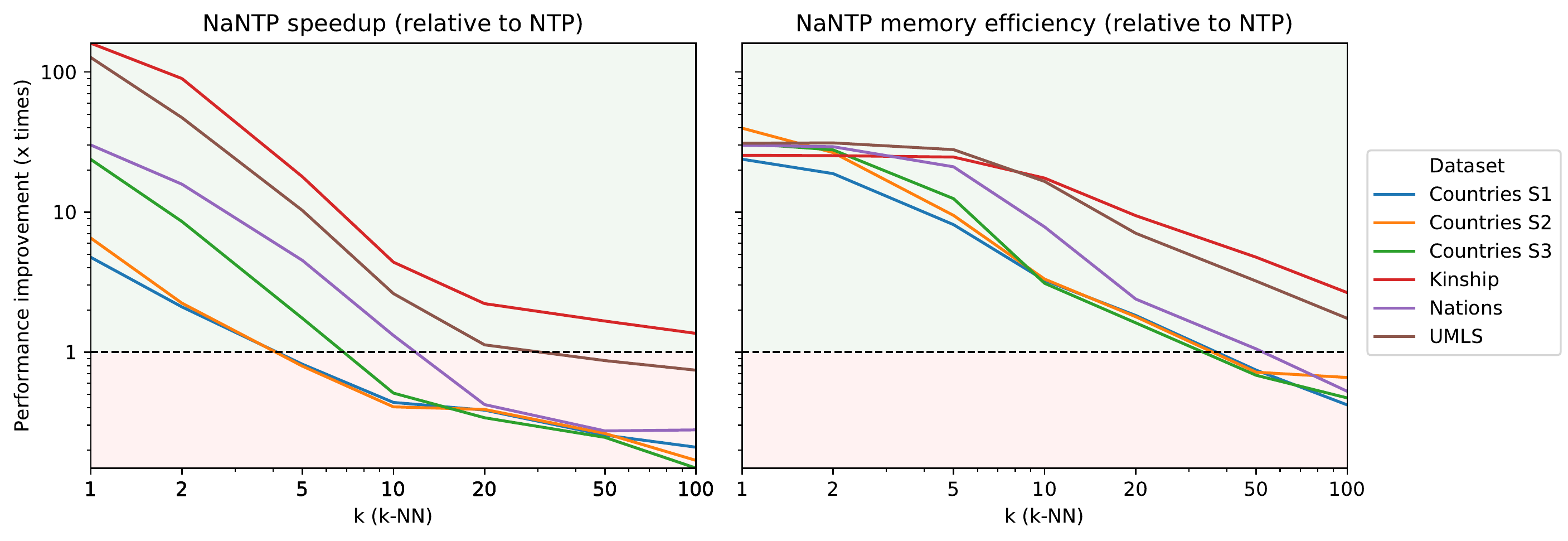}
\caption{Run-time and memory performance of \gls{NaNTP} in comparison with \gls{NTP}
Run-time speedup calculated as the ratio of examples per second of \gls{NaNTP} and \gls{NTP}. Memory efficiency calculated as a ratio of the memory use of \gls{NTP} and \gls{NaNTP}.
Dashed line denotes equal performance -- above it (green) \gls{NaNTP} performs better, below it (red) performs worse.
}
\label{fig:scalability_benchmark}
\end{figure*}

\section{Datasets} \label{app:datasets}

We run experiments on the following datasets, and report results in terms of Area Under the Precision-Recall Curve~\citep{DBLP:conf/icml/DavisG06} (AUC-PR), \gls{MRR}, and HITS@$m$~\citep{DBLP:conf/nips/BordesUGWY13}.
\subsection{Countries, UMLS, Nations} \label{app:smaller}
\subsubsection{Countries}
Countries is a dataset introduced by \citet{bouchard2015approximate} for testing reasoning capabilities of neural link prediction models.
It consists of $244$ countries, $5$ regions (\eg{} \ent{Europe}), $23$ sub-regions (\eg{} \ent{Western Europe}, \ent{North America}), and $1158$ facts about the neighbourhood of countries, and the location of countries and sub-regions.
As in \citet{rocktaschel2017end}, we randomly split countries into a training set of $204$ countries (train), a development set of $20$ countries (validation), and a test set of $20$ countries (test), such that every validation and test country has at least one neighbour in the training set.
Subsequently, three different task datasets are created, namely \textbf{S1}, \textbf{S2}, and \textbf{S3}.
For all tasks, the goal is to predict $\rel{locatedIn}(c, r)$ for every test country $c$ and all five regions $r$, but the access to training atoms in the \gls{KB} varies.
\begin{description}
    \item[S1:] All ground atoms $\rel{locatedIn}(c, r)$, where $c$ is a test country and $r$ is a region, are removed from the \gls{KB}. 
    Since information about the sub-region of test countries is still contained in the \gls{KB}, this task can be solved by using the transitivity rule:
\begin{equation*}
\begin{aligned}
\rel{locatedIn}(\var{X}, \var{Y}) \lif & \rel{locatedIn}(\var{X}, \var{Z}), \\
& \rel{locatedIn}(\var{Z}, \var{Y}).
\end{aligned}
\end{equation*}
    \item[S2:] In addition to \textbf{S1}, all ground atoms $\rel{locatedIn}(c, s)$ are removed where $c$ is a test country and $s$ is a sub-region.
    The location of countries in the test set needs to be inferred from the location of its neighbouring countries:
\begin{equation*}
\begin{aligned}
\rel{locatedIn}(\var{X}, \var{Y}) \lif & \rel{neighborOf}(\var{X}, \var{Z}), \\
& \rel{locatedIn}(\var{Z}, \var{Y}).
\end{aligned}
\end{equation*}
    This task is more difficult than \textbf{S1}, as neighbouring countries might not be in the same region, so the rule above will not always hold.
    \item[S3:] In addition to \textbf{S2}, also all ground atoms $\rel{locatedIn}(c, r)$ are removed where $r$ is a region and $c$ is a country from the training set training that has a country from the validation or test sets as a neighbour. 
    The location of test countries can for instance be inferred using the rule:
\begin{equation*}
\begin{aligned}
    \rel{locatedIn}(\var{X}, \var{Y}) \lif & \rel{neighborOf}(\var{X}, \var{Z}), \\
    & \rel{neighborOf}(\var{Z}, \var{W}), \\
    & \rel{locatedIn}(\var{W}, \var{Y}).
\end{aligned}
\end{equation*}
\end{description}

\subsubsection{Countries with Mentions} \label{app:mentions}

We generated a set of variants of Countries S1, S2, and S3, by randomly replacing a varying number of training set triples with mentions.
The employed mentions are outlined in \cref{tab:mentions}.
\subsubsection{Nations and UMLS}
Furthermore, we consider the Nations, and the \gls{UMLS} datasets~\citep{DBLP:conf/icml/KokD07}.
\gls{UMLS} contains $49$ predicates, $135$ constants and $6529$ true facts, while Nations contains $56$ binary predicates, $111$ unary predicates, $14$ constants and $2565$ true facts.
We follow the protocol used by \citet{rocktaschel2017end} and split every dataset into training, development, and test facts, with a $80\%/10\%/10\%$ ratio.
For evaluation, we take a test fact and corrupt its first and second argument in all possible ways such that the corrupted fact is not in the original \gls{KB}.
Subsequently, we predict a ranking of the test fact and its corruptions to calculate \gls{MRR} and HITS@$m$.

\subsection{WordNet and Freebase} \label{app:freebase}

We also evaluate the proposed method on WordNet (WN18) and Freebase (FB122) jointly with the set of rules released by \citet{DBLP:conf/emnlp/GuoWWWG16}.
WordNet~\citep{DBLP:journals/cacm/Miller95} is a lexical knowledge base for the English language, where entities correspond to word senses, and relationships define lexical relations between them.
The WN18 dataset consists of a subset of WordNet, containing 40,943 entities, 18 relation types, and 151,442 triples.
We also consider WN18RR~\citep{DBLP:conf/aaai/DettmersMS018}, a dataset derived from WN18 where predicting missing links is sensibly harder.
Freebase~\citep{DBLP:conf/aaai/BollackerCT07} is a large knowledge graph that stores general facts about the world.
The FB122 dataset is a subset of Freebase regarding the topics of \emph{people}, \emph{location} and \emph{sports}, and contains 9,738 entities, 122 relation types, and 112,476 triples.
For both data sets, we used the fixed training, validation, test sets and rules provided by \citet{DBLP:conf/emnlp/GuoWWWG16}; a subset of the rules is shown in \cref{tab:rules}.
Note that a subset of the test triples can be inferred by deductive logic inference.
For such a reason, following \citet{DBLP:conf/emnlp/GuoWWWG16}, we also partition the test set in two subsets, namely Test-I and Test-II: Test-I contains triples that \emph{cannot} be inferred by deductive logic inference, while Test-II contains all remaining test triples.

\section{Run-Time Performance comparison} \label{app:runtime}

To assess the run-time gains of \gls{NaNTP}, we compare it to \gls{NTP} with respect to time and memory performance during training.
In our experiments, we vary the $n$ of the \gls{NNS} to assess the computational demands by increasing $n$.
First, we compare the average number of examples (queries) per second by running 10 training batches with a maximum batch to fit the memory of NVIDIA GeForce GTX 1080 Ti, for all models.
Second, we compare the maximum memory usage of both models on a CPU, over 10 training batches with same batch sizes.
The comparison is done on a CPU to ensure that we include the size of the \gls{NNS} index in \gls{NaNTP} measures and as a fail-safe, in case the model does not fit on the GPU memory.
The results, presented in Figure \ref{fig:scalability_benchmark}, demonstrate that, compared to \gls{NTP}, \gls{NaNTP} is considerably more time and memory efficiency.
In particular, we observe that \gls{NaNTP} yields significant speedups of an order of magnitude for smaller datasets (Countries S1 and S2), and more than two orders of magnitude for larger datasets (Kinship and Nations).
Interestingly, with the increased size of the dataset, \gls{NaNTP} consistently achieves higher speedups, when compared to \gls{NTP}.
Similarly, \gls{NaNTP} is more memory efficient, with savings bigger than an order of magnitude, making them readily applicable to larger datasets, even when augmented with textual surface forms.

\section{Hyper-parameters} \label{app:hyper}

For each experiment, the best hyperparameters were selected via cross-validation.
We use Adam~\citep{DBLP:journals/corr/KingmaB14} for minimising the loss function in \cref{eq:loss}.
We searched for the best learning rates in $\{ 0.001, 0.005, 0.01, 0.05, 0.1 \}$, for the best L2 regularisation weights in $\{ 0.001, 0.0001 \}$.
For Freebase and WordNet, we fixed the batch size to 1000, while for Countries, UMLS, Kinship, and Nations we searched the best batch size in $\{ 10, 20, 50, 100 \}$.
About \glspl{NaNTP}-specific hyperparameters, we searched for the best number of rules $k_{r}$ and facts $k_{f}$ to unify with in $\{ 1, 3, 5 \}$.
Due to time and computational constraints, the embedding size of entities and relation types was set to 100, the number of epochs was also set to 100, while the maximum proof depth $d$ was fixed to 2.
In all experiments, we observed a quick convergence of the model already in the first 20-30 epochs.
On FB122, we found it useful to pre-train rules first (95 epochs), without updating any entity or relation embeddings, and then training the entity embeddings jointly with the rules (5 epochs).
This forces \glspl{NaNTP} to learn a good rule-based model of the domain before fine-tuning its representations.

\end{document}